\documentclass[10pt]{article}

\usepackage[margin=1in]{geometry}
\usepackage{amsmath,amssymb}
\usepackage{graphicx}
\usepackage{booktabs}
\usepackage{hyperref}
\usepackage{caption}
\usepackage{subcaption}
\usepackage{booktabs}
\usepackage{multirow}
\usepackage[table]{xcolor}
\usepackage{caption}
\usepackage[table]{xcolor}  % IMPORTANT: [table] option
\usepackage{colortbl}
\usepackage{adjustbox}
\usepackage{xcolor}
\usepackage{ulem} % for underline

\usepackage{tikz}
\usepackage{pgfplots}
\usepackage{subcaption}
\pgfplotsset{compat=1.18}
\usepackage{hyperref}
\usepackage{cleveref}

\usepackage{orcidlink}

\title{Learning Image-Adaptive Scale Fields for Metric Depth Recovery}
\author{Yuanyan Li, Matthias Althoff\\
Technical University of Munich \\}
\date{}

\begin{document}
\maketitle

\begin{abstract}
Monocular depth estimation (MDE) typically produces depth estimations that are defined up to an unknown scale or shift.
When only sparse metric anchors are available, recovering accurate metric depth becomes challenging yet necessary for practical applications.
We address this problem by formulating metric depth recovery as image-adaptive scale field modeling.
Instead of directly correcting the depth, we reformulate the correction as a low-dimensional linear combination of image-adaptive basis maps.
These maps are derived from semantic and geometric cues encoded in the
MDE estimations and intermediate representations.
The weights of basis maps are efficiently determined from sparse metric anchors via a least-squares problem.
This formulation yields improved metric depth accuracy, strong robustness under extreme anchor sparsity, and an interpretable decomposition of spatial scale variations.
Extensive experiments across multiple datasets and representative MDE models demonstrate the effectiveness and general applicability of our approach.
% The code and trained models will be publicly available upon acceptance of this paper.
\end{abstract}

\section{Introduction}
\label{sec:introduction}

Monocular depth estimation (MDE) has advanced significantly in recent years, 
driven by high-capacity transformer-based architectures~\cite{dpt} 
and large-scale pretraining strategies~\cite{dinov2,depthanything,lin2025depthany3}.
While several approaches aim to estimate metric depth directly~\cite{zoedepth,metric3d,depthpro}, many widely used methods still produce depth maps defined only up to unknown scale and shift~\cite{dpt,midas}. 
Moreover, metric depth estimation continues to face generalization challenges under domain shift~\cite{survey_mmde}, making explicit metric alignment essential for downstream applications, such as robotics, autonomous driving, and 3D reconstruction.
In practice, this alignment requires additional metric information, which is often only sparsely available, such as LiDAR measurements~\cite{models_sensors} or geometric constraints from structure-from-motion~\cite{Hartley_Zisserman_2004}.
These sparse metric measurements, referred to as anchors, provide absolute references for alignment.

A widely used approach aligns monocular depth estimation to anchors via a global scale-shift transformation~\cite{umeyama1991least}.
This approach is simple and stable, but assumes uniform scale and shift biases, an assumption often violated in real scenes with diverse objects, depth ranges, or complex layouts.
To relax this assumption, prior work has proposed more expressive models, including depth-dependent formulations~\cite{aamir2025robust}, locally weighted regression~\cite{xu2024toward}, grid-based interpolation~\cite{zhang2024hislam2}, and region-aware approaches~\cite{fan2025regionscale}.
While these methods improve expressiveness of the scale-shift field, they typically either rely on structural assumptions that may not hold in real-world scenes~\cite{xu2024toward,zhang2024hislam2} or fragment the anchor set for local alignment, where insufficient anchors lead to unstable estimates~\cite{aamir2025robust,fan2025regionscale}.
Existing methods differ primarily in their parameterization, imposing different priors that balance expressiveness and robustness under sparse anchor supervision.

We argue that metric depth recovery depends on how the scale-shift field is modeled and propose a low-dimensional, image-adaptive reformulation that preserves expressiveness while improving robustness under sparse supervision.
Specifically, we model the scale-shift field as a linear combination of a set of learned basis maps derived from estimated depth and intermediate representations extracted from the MDE backbone.
Sparse anchors determine the weights of basis maps through a globally aggregated least-squares problem, resulting in a stable closed-form solution even under extreme anchor sparsity.
Our formulation does not assume a specific backbone architecture and only requires intermediate representations and depth estimations.

We evaluate our approach on multiple datasets using Depth Anything 3 (DA3)~\cite{lin2025depthany3} and MiDaS~\cite{midas,dpt} as representative state-of-the-art MDE models.
Across diverse scenes and varying anchor densities, our method consistently ranks among the top-performing approaches.
Our approach offers:
(i) improved accuracy across datasets and scene types;
(ii) robustness through its low-dimensional formulation and global least-squares estimation;
and (iii) enhanced interpretability by representing scale-shift bias with a set of learned basis maps.

\section{Problem Statement and Related Work}
\subsection{Problem Statement}

Monocular depth estimation (MDE) estimates per-pixel depth from a single RGB image.
Let $D_{\mathrm{MDE}}(x)$ denote the estimated depth at pixel $x$, and let $D_{\mathrm{gt}}(x)$ denote the corresponding metric depth.
Due to inherent scale ambiguity in MDE~\cite{Eigen2014DepthMP}, $D_{\mathrm{MDE}}(x)$ is defined only up to an unknown scale.

Given a sparse anchor set $\mathcal{A} = \{ a_i \mid i = 1, \dots, N \},$
where each $a_i$ denotes a pixel at which $D_{\mathrm{gt}}(a_i)$ is available, 
the objective is to recover metric depth over the entire image from sparse anchors by estimating the scale and shift.
We adopt a general affine formulation:
\begin{equation}
D_{\mathrm{gt}}(x) = s(x)\,D_{\mathrm{MDE}}(x) + t(x),
\label{eq:scale_shift}
\end{equation}
where $s(x)$ and $t(x)$ denote unknown scale and shift fields to be estimated.
While $t(x)$ is not strictly necessary under certain parameterizations, 
we include it for generality so that \cref{eq:scale_shift} provides a common template 
covering the related methods reviewed later.

\subsection{Related Work}
We review representative methods for metric depth recovery from sparse anchors.

\noindent\textbf{Depth-only Affine Alignment.}
A common class of methods assumes constant scale and shift in \cref{eq:scale_shift}, and estimates these parameters by fitting the affine model to sparse anchors via least-squares regression~\cite{umeyama1991least}:
\begin{equation}
(s^*, t^*) = 
\arg\min_{s,\,t}
\sum_{x \in \mathcal{A}}
\left(
D_{\mathrm{gt}}(x) - \big( s\,D_{\mathrm{MDE}}(x) + t \big)
\right)^2.
\label{eq:global_scale}
\end{equation}
While numerically stable, non-uniform scale and shift biases cannot be captured.

Piecewise affine alignment~\cite{aamir2025robust} extends the model in \cref{eq:global_scale} by estimating separate scale and shift parameters over predefined depth intervals.
Although this relaxes the global assumption by allowing interval-specific $s$ and $t$, it fragments the anchor set across intervals, resulting in fewer anchors per interval and thus reduced robustness.
Moreover, \cite{aamir2025robust} operates purely in depth space and cannot model spatial variation across the image.

\noindent\textbf{Locally Weighted Regression.}
Locally weighted linear regression (LWLR)~\cite{xu2024toward} estimates spatially varying scale and shift parameters in \cref{eq:scale_shift}. 
By solving \cref{eq:global_scale}, a globally aligned depth $\tilde{D}(x)$ is first computed, followed by a local regression.
For a pixel $x$, local parameters $\boldsymbol{\beta}_{x} = [s_{x},\, t_{x}]^\top$ are obtained by minimizing the following objective~\cite[Eq.~(5)]{xu2024toward}:
\begin{equation}
(\mathbf{y}-\mathbf{X}\boldsymbol{\beta}_{x})^\top
\mathbf{W}_{x}
(\mathbf{y}-\mathbf{X}\boldsymbol{\beta}_{x})
+
\lambda t_{x}^2,
\end{equation}
where $\mathbf{y}$ contains the anchor depth measurements and 
$\mathbf{X} = [\tilde{D},\, 1]$ stacks $\tilde{D}(x)$ at anchor pixels together with a column of ones corresponding to $t_{x}$.
Since $\mathbf{W}_x$ assigns larger weights to anchors near $x$, the fitted parameters $\boldsymbol{\beta}_x$ mainly reflect the local scale-shift relationship supported by nearby anchors.
This effectively acts as a low-pass spatial prior, enforcing smooth spatial variation of scale and shift, which may be violated near depth discontinuities or semantic boundaries.

\noindent\textbf{Grid-based Scale Optimization.}
Grid-based methods~\cite{zhang2024hislam2} parameterize a spatially varying scale field on a coarse grid.
Each grid vertex is associated with an unknown scale coefficient $s_k$.
For a pixel $x$, let $\mathcal{N}(x)$ denote its four neighboring grid vertices.
The scale $s(x)$ in \cref{eq:scale_shift} is obtained by bilinear interpolation:
\begin{equation}
s(x)=\sum_{k\in\mathcal{N}(x)} w_k(x)\, s_k,
\quad
\sum_{k\in\mathcal{N}(x)} w_k(x)=1,
\end{equation}
where $w_k(x)$ are bilinear weights determined by the relative position of $x$ inside the grid cell.
% This can be viewed as a special case of \cref{eq:scale_shift} with $t(x)=0$.
The scale field is constrained to be a bilinear interpolation of $s_k$, limiting expressiveness for spatial scale variations.
Moreover, the fixed grid is agnostic to image content, preventing adaptation to scene geometry.

\noindent\textbf{Region-aware Scale Adaptation.}
Region-aware methods~\cite{fan2025regionscale} use semantic or instance-level segmentation to partition the image into disjoint regions $R_i$ and estimate scale and shift independently within each $R_i$.
This can be interpreted as a region-wise parameterization of \cref{eq:scale_shift}.
Specifically, for $x\in R_i$, the metric depth at anchor pixels is modeled as
\begin{equation}
D_{\mathrm{gt}}(x)
=
s_i\, D_{\mathrm{MDE}}(x)
+
\beta_i\, u
+
\gamma_i\, v
+
t_i,
\end{equation}
where $(u,v)$ are coordinates of $x$, and $(s_i,\beta_i,\gamma_i,t_i)$ are region-specific parameters.
This formulation can be interpreted as a special case of Eq.~\eqref{eq:scale_shift}, i.e., for $x \in R_i$,
$s(x) = s_i$ and
$t(x) = \beta_i u + \gamma_i v + t_i.$
The fitted model is then applied to all pixels in the region, with regions iteratively expanded when anchors are insufficient.

Although region-wise parameterization improves expressiveness, it is generally less robust than \cref{eq:global_scale} due to the limited anchors per region.
Moreover, reliance on external segmentation models~\cite{oneformer,segment_any} increases deployment complexity and may propagate segmentation errors during  alignment, potentially misaligning region boundaries with actual object boundaries or depth discontinuities.

In general, recovering dense per-pixel scale and shift fields from sparse anchors is inherently ill-posed, since sparse anchors are insufficient to uniquely determine varying scale and shift at every pixel. 
Existing methods address this challenge by imposing predefined structural priors on $s(x)$ and $t(x)$; 
however, such assumptions may not hold in real-world scenes and can become unreliable under sparse supervision.

% \section{Problem Formulation}
% \input{sections/problem_formulation}

\section{Method}
\subsection{Parameterization of the Scale Field}
\label{subsec:method_explain}

\begin{figure*}[t]
\centering
\includegraphics[width=\textwidth]{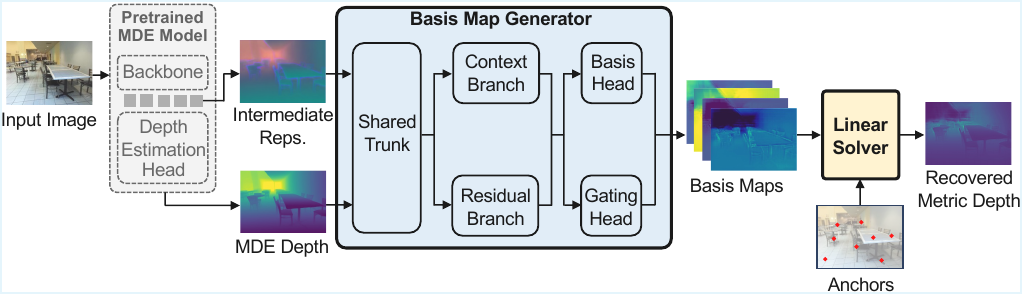}
\caption{
Overview of our approach.
A pretrained monocular depth estimation (MDE) model produces an initial depth estimation $D_{\mathrm{MDE}}(x)$ and intermediate representations $F(x)$.
The basis map generator takes these outputs and produces a $K$-dimensional embedding $E(x)=[E_1(x),\dots,E_K(x)]$, whose components define $K$ basis maps.
Sparse metric anchors are used to weight these maps via a least-squares problem to obtain a spatially varying scale field.
}
\label{fig:method_overview}
\end{figure*}

We parameterize the scale field $s(x)$ in \cref{eq:scale_shift} by weighting basis maps derived from semantic and geometric cues encoded in the estimated depth as well as intermediate representations extracted from the MDE backbone.
An overview of our approach is illustrated in \cref{fig:method_overview}. 

Instead of modeling the scale field directly, we operate in log space and define
\begin{equation}
\label{eq:log_scale}
\ell(x) = \log s(x).
\end{equation}
When modeling $s(x)$ directly, the aligned depth is given by $s(x) D_{\mathrm{MDE}}(x)$, and the gradient with respect to $s(x)$ satisfies
\begin{equation}
\frac{\partial}{\partial s(x)} \big(s(x) D_{\mathrm{MDE}}(x)\big) = D_{\mathrm{MDE}}(x).
\end{equation}
Thus, pixels with larger depth values induce larger gradient magnitudes, leading to depth-dependent optimization where large-depth pixels dominate the updates.
By parameterizing in log space with \cref{eq:log_scale}, the multiplicative scale becomes additive in $\ell(x)$, mitigating this depth-dependent gradient scaling.

We adopt a low-dimensional parameterization for $\ell(x)$.
A neural network, termed the basis map generator, predicts a $K$-dimensional embedding
$
E(x) \in \mathbb{R}^{K}.
$
Accordingly, the log-scale field is represented as
\begin{equation}
\label{eq:scale_para}
\ell(x) = E(x)^{\top} \mathbf{w} = \sum_{m=1}^K w_m E_m(x),
\end{equation}
where $\mathbf{w}\in\mathbb{R}^K$ are image-specific weights, and each component $E_m(x)$ defines a scalar field over the image; these fields serve as $K$ image-adaptive basis maps whose linear combination represents $\ell(x)$.
The recovered metric depth is
\begin{equation}
\label{eq:estimate_depth}
\hat{D}(x)
=
D_{\mathrm{MDE}}(x)\, e^{\ell(x)}.
\end{equation}
At anchors $a_i$, \cref{eq:scale_para} predicts 
$
\ell(a_i)=E(a_i)^\top\mathbf{w}.
$
The corresponding target value is
\begin{equation}
\label{eq:anchor_targets}
y_i
=
\log D_{\mathrm{gt}}(a_i)
-
\log D_{\mathrm{MDE}}(a_i).
\end{equation}
We estimate $\mathbf{w}$ by least-squares fitting, minimizing the squared residuals
\begin{equation}
\sum_{i=1}^N (E(a_i)^\top \mathbf{w}-y_i)^2 .
\end{equation}
Defining
\begin{equation}
\label{eq:M}
M_{i,m}=E_m(a_i),
\end{equation}
the $i$-th residual becomes
\begin{equation}
E(a_i)^\top \mathbf{w}-y_i
=
\sum_{m=1}^K M_{i,m} w_m - y_i,
\end{equation}
Stacking all anchors yields the vector form $M\mathbf{w}-\mathbf{y}$.
We solve the regularized least-squares problem to stabilize the estimation under sparse anchors
\begin{equation}
\label{eq:ridge_regression}
\mathbf{w}^*
=
\arg\min_{\mathbf{w}}
\| M\mathbf{w} - \mathbf{y} \|_2^2
+
\lambda \|\mathbf{w}\|_2^2.
\end{equation}
with closed-form solution~\cite[Eq.~(3.44)]{Hastie+al:2009}
\begin{equation}
\label{eq:closed_form_solution}
\mathbf{w}^*
=
(M^\top M + \lambda I)^{-1} M^\top \mathbf{y}.
\end{equation}
Substituting \cref{eq:closed_form_solution} into \cref{eq:estimate_depth} yields the dense estimation
\begin{equation}
\hat{D}(x)
=
D_{\mathrm{MDE}}(x)\, e^{E(x)^\top \mathbf{w}^*}.
\end{equation}
Thus, inference reduces to a small regression problem in an image-adaptive space: sparse anchors determine the weights $\mathbf{w}$, while the learned basis maps specify the representation of the scale field.

\subsection{Basis Map Formulation}
\label{subsec:basis_formulation}

We next describe how the basis maps are generated.

\begin{figure}[t]
\centering
\includegraphics[width=0.8\linewidth]{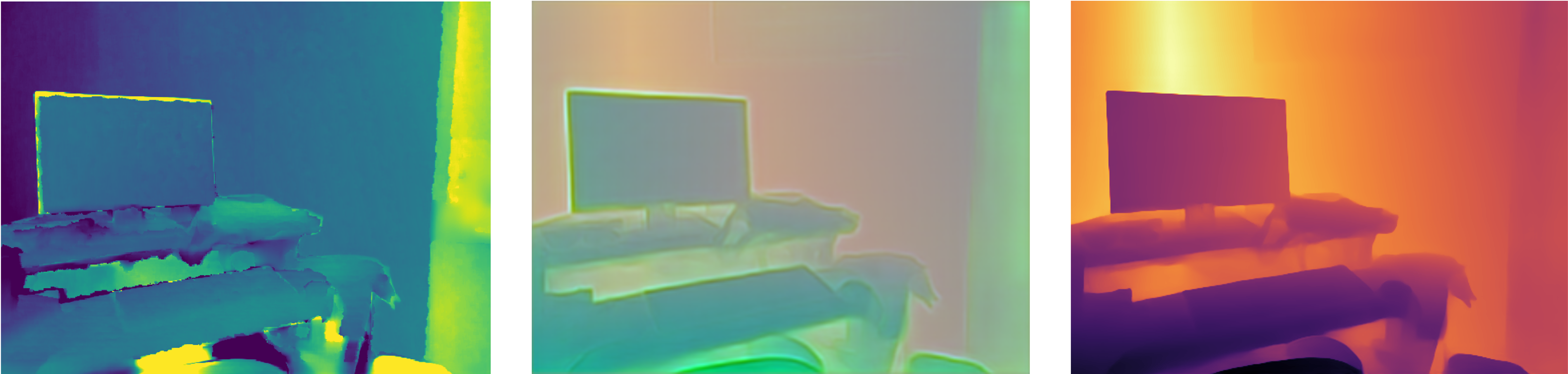}
\caption{
Example from SUN RGB-D~\cite{Song2015SUNRA} with DA3~\cite{lin2025depthany3}.
Left to right: scale field, intermediate representation, and MDE depth.
Their spatial correspondence highlights how semantic and geometric cues in the MDE outputs relate to scale variation.
}

\label{fig:justification_visualization}
\end{figure}

\noindent\textbf{Cue-conditioned design.}
Scale bias in MDE can exhibit different forms, ranging from global offsets to spatially structured errors.
To model such variability, the basis maps are generated from geometric, semantic, and spatial cues. Specifically, the basis map generator takes as input (i) intermediate representations $F(x)$ extracted from the MDE backbone prior to the depth estimation head, (ii) the estimated log-depth $\log D_{\mathrm{MDE}}(x)$, and (iii) image coordinates $x$. These values provide complementary information: $F(x)$ captures semantic and structural context, $\log D_{\mathrm{MDE}}(x)$ encodes depth-dependent geometry together with scene layout cues, and $x$ enables the modeling of spatial trends.
\Cref{fig:justification_visualization} qualitatively illustrates this relationship: the scale field exhibits strong spatial alignment with both the intermediate representations and the MDE depth estimation, supporting the use of cue-conditioned basis maps for modeling scale variation.

\noindent\textbf{Architecture.}
Let
$
X(x)
=
\bigl[
F(x),\,
\log D_{\mathrm{MDE}}(x),\,
\nabla \log D_{\mathrm{MDE}}(x),\,
x
\bigr]
$denote the input to the basis map generator, where 
$\nabla \log D_{\mathrm{MDE}}(x)$ is the spatial gradient of $\log D_{\mathrm{MDE}}(x)$, capturing local depth discontinuities and geometric transitions.

A fully convolutional trunk $\mathcal{T}_\theta$ produces shared features 
$H(x)=\mathcal{T}_\theta(X(x))$.
From $H(x)$, a basis head predicts primitive basis maps
\begin{equation}
B(x)=[B_0(x),\dots,B_{K-1}(x)],
\end{equation}
with $B_0(x)\equiv 1$, introducing a global scale mode so that global alignment is representable.
The remaining primitive basis maps are learned from data through the basis head and therefore adapt to spatial bias patterns present in the training data, allowing localized deviations and spatially varying adjustments.

A gating head predicts a $K$-channel map $G(x)$.
The gating weights are normalized by a softmax so that, at each pixel,
\begin{equation}
\label{eq:gating_simplex}
G_m(x)\ge0,
\qquad
\sum_{m=0}^{K-1} G_m(x)=1.
\end{equation}
The final basis maps are obtained via element-wise multiplication
\begin{equation}
\label{eq:element_modu}
E(x)=G(x)\odot B(x), \qquad E_m(x)=G_m(x)B_m(x).
\end{equation}

Under the parameterization in \cref{eq:scale_para}, $\ell(x)$ is expressed as a linear combination of basis maps with image-level weights $w_m$.
Since each weight $w_m$ is spatially invariant, a single
channel $E_m(x)$ can only be scaled uniformly over all pixels.
As a result, if a spatial pattern needs to contribute to $\ell(x)$
with different magnitudes in different regions of the image,
this variation cannot be represented within a single channel.
Consequently, if the basis maps $E_m(x)$ are predicted directly,
when the same spatial pattern is required in multiple regions with different magnitudes, it must be duplicated across several channels.

To avoid this redundancy, we factorize $E_m(x)$ as in
\cref{eq:element_modu}.
The primitive maps $B_m(x)$ represent spatial components of the scale
field that are shared across regions, while the gating maps $G_m(x)$ determine where and how strongly each component contributes to $\ell(x)$.
This separation allows the same spatial pattern $B_m(x)$ to be reused
in different parts of the image, with its contribution controlled
by $G_m(x)$.

The softmax constraint in \cref{eq:gating_simplex} ensures that the
gating maps form a normalized mixture at each pixel.
Therefore, they cannot arbitrarily scale the magnitude of each component.
The overall magnitude is determined by the weights
$\mathbf{w}$ in \cref{eq:scale_para}, while $G_m(x)$ controls their
relative contribution across spatial locations.
Under this formulation, $\ell(x)$ remains linear in $\mathbf{w}$ while
allowing spatially varying modulation through $G_m(x)$.

\noindent\textbf{Generalization to existing models.}
Our approach subsumes several existing alignment methods~\cite{umeyama1991least,aamir2025robust,xu2024toward,zhang2024hislam2,fan2025regionscale} as special cases.
Consider affine alignment models \cref{eq:scale_shift}, 
which explicitly introduce a shift term $t(x)$.
This model can be equivalently rewritten as
\begin{equation}
\label{eq:absorb_shift}
\ell^*(x)
=
\log\!\left(
s(x) + \frac{t(x)}{D_{\mathrm{MDE}}(x)}
\right),
\end{equation}
so that the shift term appears as a depth-dependent correction of the scale.
Since $D_{\mathrm{MDE}}(x)$ is an input to the basis map generator,
such depth-dependent corrections can be modeled by the learned scale field,
without introducing an explicit shift component.

Furthermore, the inclusion of spatial coordinates $x$ enables spatially varying corrections that can approximate locally weighted~\cite{xu2024toward} or grid-based~\cite{zhang2024hislam2} adjustments, while conditioning on $F(x)$ allows the model to capture region-consistent scale behavior similar to region-wise decomposition approaches~\cite{fan2025regionscale}, without requiring explicit region partitioning.

\subsection{Spectral Structure and Energy Concentration}
\label{subsec:kernel}
Since the log-scale field is parameterized linearly in \cref{eq:scale_para},
each anchor $a_i$ is associated with a $K$-dimensional embedding vector
$E(a_i)\in\mathbb{R}^K$.
For an anchor set $\mathcal A$, 
let $M\in\mathbb R^{N\times K}$ denote the embedding matrix
defined in \cref{eq:M}.
The associated Gram matrix
\begin{equation}
C = M M^\top,
\qquad
C_{ij}=E(a_i)^\top E(a_j),
\label{eq:gram_matrix}
\end{equation}
measures pairwise inner products between anchor embedding vectors.
Let its eigendecomposition be
\begin{equation}
C = U\,\mathrm{diag}(s_1,\dots,s_N)\,U^\top.
\label{eq:gram_decompo}
\end{equation}

The eigenvalues $\{s_\ell\}$ characterize how the anchor embeddings
$\{E(a_i)\}$ are distributed across directions in $\mathbb R^K$:
large eigenvalues correspond to directions along which many anchors
exhibit coherent variation, while small eigenvalues indicate directions
that are weakly supported by the data~\cite{prml}.

Let $\mathbf y=(y_1,\dots,y_N)^\top\in\mathbb R^N$ denote the anchor
target values defined in \cref{eq:anchor_targets}.
Expanding $\mathbf y$ in the eigenbasis of $C$,
\begin{equation}
\mathbf y=\sum_{\ell=1}^{N}\alpha_\ell u_\ell,
\label{eq:anchor_in_eigenbasis}
\end{equation}
where $u_\ell$ are the eigenvectors of $C$, decomposes $\mathbf y$ into components along the eigenvectors of $C$.

As global scale bias affects pixels coherently and is often
the dominant component of bias~\cite{Eigen2014DepthMP},
the anchor embeddings $E(a_i)$ tend to align along a common direction,
leading to a dominant eigenvalue $s_1$.
Consequently, the anchor signal $\mathbf y$ is expected to concentrate
on the leading eigenmode, so that $|\alpha_1|$ dominates
$|\alpha_\ell|$ for $\ell>1$.
Since directions associated with large eigenvalues are well supported by $E(a_i)$, the dominant component can be reliably estimated even from few anchors.
Additional anchors mainly improve estimation along weakly supported directions corresponding to smaller eigenvalues.

\subsection{Training Objective}

To train the basis map generator, we optimize it using the following loss terms.
\noindent\textbf{Dense Loss.}
We supervise the reconstructed metric depth using:
\begin{equation}
\mathcal{L}_{\text{dense}} =
\|\log \hat{D}(x)-\log D_{\text{gt}}(x)\|_{\text{SmoothL1}}.
\end{equation}
This term encourages accurate estimation of the spatially varying scale field, while the SmoothL1 norm provides robustness to outliers~\cite{smoothl1}.

\noindent\textbf{Anchor Loss.}
At anchor pixels, we enforce consistency between the estimated scale field and the sparse metric constraints:
\begin{equation}
\mathcal{L}_{\text{anchor}} =
\frac{1}{N}\sum_{i=1}^{N}\left((M\mathbf{w}^*)_i-y_i\right)^2,
\end{equation}
so that the estimated scale field matches the anchor measurements, providing direct supervision.

\noindent\textbf{Regularization.}
To encourage diversity among basis maps, we introduce a decorrelation loss. 
Let $\tilde{e}_i$ denote the normalized vector of the $i$-th basis map, based on which we define
\begin{equation}
\mathcal{L}_{\text{decor}} =
\frac{1}{K(K-1)}\sum_{i\neq j}(\tilde{e}_i^\top \tilde{e}_j)^2,
\end{equation}
favoring orthogonality among basis maps and preventing degenerate or redundant modes in the learned subspace.
We further regularize the gating maps via negative entropy,
\begin{equation}
\mathcal{L}_{\text{gate}} =
\sum_x\sum_m G_m(x)\log G_m(x),
\end{equation}
which encourages higher-entropy gating distributions and discourages overly sharp, one-hot assignments.

\noindent\textbf{Overall Loss.}
The overall loss is
\begin{equation}
\mathcal{L}=
\mathcal{L}_{\text{dense}}
+\lambda_{\text{anchor}}\mathcal{L}_{\text{anchor}}
+\lambda_{\text{decor}}\mathcal{L}_{\text{decor}}
+\lambda_{\text{gate}}\mathcal{L}_{\text{gate}}.
\end{equation}

Overall, this training objective provides the necessary supervision and regularization to realize the proposed formulation under sparse anchor constraints.

\section{Experiments}
\label{experiments}

\subsection{Experimental Setup}

All experiments are conducted on recorded data using frozen MDE models. We use Depth Anything 3 (DA3-BASE)~\cite{lin2025depthany3} and MiDaS (DPT-Hybrid-384)~\cite{midas,dpt} to obtain MDE depth and intermediate representations.

\noindent\textbf{Datasets.}
All methods follow strict train/val/test splits, with the test set used exclusively for evaluation.
Experiments are conducted on KITTI Raw dataset~\cite{Geiger2013IJRR}, UrbanSyn~\cite{GOMEZ2025130038}, and SUN RGB-D~\cite{Song2015SUNRA}, reporting AbsRel and $\delta_1$ within a predefined depth mask.
For brevity, we refer to KITTI Raw as KITTI in the remainder of this paper.
The depth mask excludes sky regions, invalid pixels, depths $<0.1\,\mathrm{m}$, and depths exceeding $10\,\mathrm{m}$ (SUN RGB-D) or $80\,\mathrm{m}$ (KITTI, UrbanSyn)~\cite{bhat2021adabins,umde}.
KITTI uses 1009/100/289 frames for train/val/test with video-level splits across \emph{City}, \emph{Residential}, \emph{Road}, and \emph{Campus}. 
SUN RGB-D uses 8267/1034/1034 images, and UrbanSyn uses 6031/754/754 images.

\noindent\textbf{Anchor Extraction.}
Sparse metric anchors are sampled per image within the depth mask.
A fixed $A\times B$ grid defines candidate anchor pixels; candidates outside the mask are shifted to the nearest valid pixel.
We evaluate three anchor regimes: low ($N\!\in\![10,15]$), medium ($N\!\in\![100,120]$), and high ($N\!\in\![500,530]$), with $N$ randomly sampled within each interval (fixed seed).

\noindent\textbf{Implementation Details.}
Global~\cite{umeyama1991least}, piecewise~\cite{aamir2025robust}, LWLR~\cite{xu2024toward}, grid-based alignment~\cite{zhang2024hislam2} and region-aware method~\cite{fan2025regionscale} are re-implemented by us, as no official evaluation code is publicly available. 
All methods are implemented strictly following the descriptions in the original papers, with hyperparameters set according to the reported configurations whenever specified. 
The region-aware method~\cite{fan2025regionscale} is evaluated on SUN RGB-D and UrbanSyn using ground-truth semantic masks for alignment to avoid introducing additional segmentation errors.
It is not evaluated on KITTI, as the dataset does not provide ground-truth semantic or instance annotations required by the method.

For our method, separate models are trained for each MDE model using a shared architecture, training protocol, and identical optimization hyperparameters to ensure a fair comparison. The loss weights are fixed across all experiments as
$\lambda_{\text{anchor}} = 0.1$ and $\lambda_{\text{decor}} = \lambda_{\text{gate}} = 1\times10^{-4}$, without per-dataset tuning.
We use $K\in\{8,30,50\}$ basis maps. 
The ridge regularization coefficient in \cref{eq:ridge_regression} is set to a small constant on the order of $10^{-3}$ to stabilize the least-squares solver.
The $K=8$ model is trained under the low-anchor regime for 25 epochs and evaluated under low, medium, and high anchor regimes, while the $K\in\{30,50\}$ models are trained under the medium-anchor regime for 25 epochs and evaluated under medium and high anchor regimes. 
The high-anchor regime is not used for initial training. 
When evaluated under an anchor regime different from its training regime, each $K$-specific model is fine-tuned for 5 epochs on the training split to ensure stable performance across varying anchor densities. 
For each run, the checkpoint with the lowest validation loss is selected. 
The proposed method relies solely on the MDE depth estimation and its intermediate representations, without using any external models or additional ground-truths.

\noindent\textbf{Drop-anchor Robustness Test.}
To evaluate extreme sparsity, we introduce a drop-anchor protocol. Starting from $9$ anchors, we iteratively remove the anchor closest to any other anchor, producing increasingly sparse sets. 
The procedure is deterministic per image and is evaluated using the $K=8$ models.

\noindent\textbf{Ablation Study.}
We analyze the role of input cues on DA3 and MiDaS by comparing the full model with a variant where $F(x)=0$ under the medium anchor regime ($N\!\in\![100,120]$), while varying $K$ to study the interaction between basis capacity (i.e., the number of basis maps) and input cues.

In the ablation study, the network architecture and all optimization settings remain unchanged.
The intermediate representation $F(x)$ is set to zero, disabling this pathway without altering model capacity or introducing additional parameters.
All other inputs and training conditions are kept identical, isolating the contribution of $F(x)$.

\subsection{Results}

\noindent\textbf{Standard Benchmark Results.}
\Cref{tab:kitti_results,tab:us_results} report results on KITTI and the combined UrbanSyn–SUN RGB-D benchmark using two MDE models under three anchor regimes. Results are shown separately, as the region-aware method is not evaluated on KITTI due to the absence of required ground-truth semantic or instance annotations.

% kitti
\begin{table*}[t]
\centering
\caption{KITTI results with different anchor numbers.
Low/Med/High correspond to 10–15 / 100–120 / 500–530 anchors per image.
Best results are in \textbf{bold}, second-best are \underline{underlined}.}
\label{tab:kitti_results}

\footnotesize
\setlength{\tabcolsep}{6pt}
\renewcommand{\arraystretch}{1.2}

\resizebox{\textwidth}{!}{%
\begin{tabular}{l cc cc cc cc cc cc}
\toprule
& \multicolumn{6}{c}{DA3} & \multicolumn{6}{c}{MiDaS} \\
\cmidrule(lr){2-7}\cmidrule(lr){8-13}

Method
& \multicolumn{2}{c}{Low}
& \multicolumn{2}{c}{Med}
& \multicolumn{2}{c}{High}
& \multicolumn{2}{c}{Low}
& \multicolumn{2}{c}{Med}
& \multicolumn{2}{c}{High} \\

& AbsRel & $\delta_1$
& AbsRel & $\delta_1$
& AbsRel & $\delta_1$
& AbsRel & $\delta_1$
& AbsRel & $\delta_1$
& AbsRel & $\delta_1$ \\
\midrule

Global
& 0.31 & 0.53
& 0.21 & 0.65
& 0.20 & 0.69

& \underline{0.17} & \underline{0.75}
& 0.16 & 0.76
& 0.16 & 0.77 \\

Grid
& \underline{0.24} & 0.63
& 0.17 & 0.77
& 0.16 & 0.78
& 0.23 & 0.72
& 0.18 & 0.79
& 0.17 & 0.80 \\

LWLR
& 0.26 & 0.63
& 0.19 & 0.69
& 0.19 & 0.70

& 0.27 & 0.56
& 0.12 & 0.82
& 0.12 & 0.84 \\

Piecewise
& 0.25 & \underline{0.66}
& 0.17 & 0.76
& 0.17 & 0.77
& 0.25 & 0.65
& 0.16 & 0.75
& 0.16 & 0.75 \\

\midrule

Basis
& \textbf{0.15} & \textbf{0.81}
& 0.13 & 0.85
& 0.12 & 0.85

& \textbf{0.13} & \textbf{0.83}
& 0.12 & 0.86
& 0.11 & 0.87 \\

Basis*
& -- & --
& \underline{0.11} & \underline{0.87}
& \underline{0.10} & \underline{0.88}

& -- & --
& \underline{0.10} & \underline{0.88}
& \underline{0.09} & \underline{0.90} \\

Basis**
& -- & --
& \textbf{0.10} & \textbf{0.88}
& \textbf{0.10} & \textbf{0.88}

& -- & --
& \textbf{0.09} & \textbf{0.89}
& \textbf{0.09} & \textbf{0.90} \\

\bottomrule
\end{tabular}}
\end{table*}

% urbansyn & sunrgbd
\begin{table*}[t]
\centering
\caption{Weighted-average results over UrbanSyn and SUN RGB-D (weighted by image count).
Low/Med/High correspond to 10–15 / 100–120 / 500–530 anchors per image.
Best results are in \textbf{bold}, second-best are \underline{underlined}.}
\label{tab:us_results}

\footnotesize
\setlength{\tabcolsep}{6pt}
\renewcommand{\arraystretch}{1.2}

\resizebox{\textwidth}{!}{%
\begin{tabular}{l cc cc cc cc cc cc}
\toprule
& \multicolumn{6}{c}{DA3} & \multicolumn{6}{c}{MiDaS} \\
\cmidrule(lr){2-7}\cmidrule(lr){8-13}

Method
& \multicolumn{2}{c}{Low}
& \multicolumn{2}{c}{Med}
& \multicolumn{2}{c}{High}
& \multicolumn{2}{c}{Low}
& \multicolumn{2}{c}{Med}
& \multicolumn{2}{c}{High} \\

& AbsRel & $\delta_1$
& AbsRel & $\delta_1$
& AbsRel & $\delta_1$
& AbsRel & $\delta_1$
& AbsRel & $\delta_1$
& AbsRel & $\delta_1$ \\
\midrule

Global
& 0.21 & 0.74
& 0.18 & 0.77
& 0.18 & 0.78

& 0.17 & 0.82
& 0.15 & 0.83
& 0.13 & 0.84 \\

Grid
& 0.29 & 0.70
& 0.22 & 0.74
& 0.22 & 0.74

& 0.38 & 0.68
& 0.29 & 0.73
& 0.29 & 0.74 \\

LWLR
& 0.19 & 0.76
& 0.12 & 0.87
& 0.12 & 0.87

& \underline{0.12} & \underline{0.85}
& \textbf{0.07} & 0.92
& \underline{0.09} & 0.91 \\

Piecewise
& \underline{0.16} & \underline{0.83}
& 0.12 & 0.86
& 0.11 & 0.88

& 0.23 & 0.78
& 0.12 & 0.86
& 0.12 & 0.86 \\

Region
& 0.32 & 0.74
& 0.09 & 0.91
& 0.06 & 0.95

& 0.22 & 0.80
& \underline{0.09} & 0.93
& \textbf{0.06} & \textbf{0.95} \\

\midrule

Basis
& \textbf{0.08} & \textbf{0.92}
& 0.07 & 0.95
& 0.07 & 0.95

& \textbf{0.14} & \textbf{0.89}
& 0.10 & \textbf{0.95}
& 0.11 & 0.92 \\

Basis*
& -- & --
& \underline{0.06} & \underline{0.95}
& \underline{0.05} & \underline{0.96}

& -- & --
& 0.10 & 0.93
& 0.09 & 0.94 \\

Basis**
& -- & --
& \textbf{0.06} & \textbf{0.95}
& \textbf{0.05} & \textbf{0.96}

& -- & --
& 0.10 & \underline{0.94}
& 0.09 & \underline{0.94} \\

\bottomrule
\end{tabular}}
\end{table*}

Reported methods correspond to different scale-alignment strategies: \emph{Global} (global affine alignment), \emph{Piecewise} (piecewise affine alignment), \emph{LWLR} (locally weighted linear regression), \emph{Grid} (grid-based optimization), and \emph{Region} (region-aware scale adaptation). Our method uses $8$ basis maps (\emph{Basis}), with higher-capacity variants \emph{Basis$^\ast$} and \emph{Basis$^{\ast\ast}$} using $30$ and $50$ basis maps.

Across datasets and anchor regimes, our approach demonstrates consistently good performance. Under low-anchor settings, it clearly outperforms global and piecewise alignment, indicating that spatially varying bias cannot be adequately captured by depth-only affine models. Unlike region-based methods that partition anchors into disjoint subsets, our approach uses all anchors to compute weights, improving robustness under sparse anchors. The learned basis maps also capture long-range bias correlations that are difficult for local schemes, such as LWLR or region-limited approaches.

As anchor density increases, higher-capacity variants remain competitive.
% Rather than committing to a fixed structural prior, our approach captures multiple modes of bias variation and adaptively combines them per image.
Increasing $K$ yields lower AbsRel and higher $\delta_1$ across regimes, showing that higher basis capacity improves modeling expressiveness and depth recovery accuracy.
Overall, the results demonstrate that our method achieves strong and consistent performance across diverse anchor regimes and MDE models.

\noindent\textbf{Drop-anchor Robustness.}
\Cref{tab:drop_anchor_results} reports results for $N\in\{1,3,5,7,9\}$ anchors, averaged across three datasets.
While performance degrades as anchors are removed, our method remains stable even under extreme sparsity.

To analyze this behavior, we examine the spectral structure of the Gram matrix.
All spectral statistics are computed using the $K=8$ model under the low-anchor regime, with DA3 as the underlying MDE model.
The learned basis maps define the Gram matrix $C$ in \cref{eq:gram_matrix},
whose eigen-decomposition (\cref{eq:gram_decompo}) characterizes the dominant directions of variation in the anchor embeddings $\{E(a_i)\}$.
We quantify the concentration of eigenvalues using
\begin{equation}
\eta_1
=
\frac{s_1}{\sum_{\ell=1}^{N} s_\ell},
\end{equation}
where $\{s_\ell\}$ are the eigenvalues of $C$.
Across datasets, the median $\eta_1$ ranges from $0.92$ to $0.97$, indicating that the energy is strongly concentrated in a single principal mode.
Since $C = MM^\top$ and $M^\top M$ share identical non-zero eigenvalues,
$\eta_1$ also reflects the spectral dominance of $M^\top M$.

We next investigate whether this dominant spectral component corresponds to the global scale bias. Consider the eigen-decomposition of $M^\top M \in \mathbb{R}^{K \times K}$:
\begin{equation}
M^\top M v_1 = \mu_1 v_1,
\end{equation}
where $v_1 \in \mathbb{R}^K$ is the leading eigenvector.
Because $M^\top M$ captures correlations among the basis channels,
its leading eigenvector $v_1$ identifies the dominant linear combination of basis maps.
Writing
\begin{equation}
v_1 = (v_{1,0}, \dots, v_{1,K-1})^\top,
\end{equation}
each entry $v_{1,k}$ indicates how strongly basis channel $E_k$ participates in the dominant eigen-direction.
Moreover, since $A = M^\top M + \lambda I$ shifts eigenvalues without affecting eigenvectors, its leading eigenvector coincides with that of $M^\top M$.
To quantify the contribution of global basis channel $E_0$,
we introduce the weight vector
\begin{equation}
g = (1,0,\dots,0)^\top \in \mathbb{R}^K,
\end{equation}
which activates only $E_0$.
We measure the similarity
\begin{equation}
\mathrm{similarity}
=
\left| v_1^\top g \right|
=
|v_{1,0}|,
\end{equation}
which equals the magnitude of the weight of $E_0$ in the dominant eigenvector.
Empirically, this value is consistently high
(median $\approx 0.991$ on KITTI,
$0.985$ on UrbanSyn, and $0.991$ on SUN RGB-D),
indicating that the dominant eigenvector is almost entirely concentrated on the global basis channel.
We also examine the spatial activation $G_0$ of this channel.
Across all datasets, the dataset-level statistics
$\mathbb{E}[G_0]\approx 0.21$ and $\mathrm{Var}(G_0)\approx10^{-4}$
indicate that the global channel is broadly activated across the image
rather than concentrated in localized regions.

The spectral dominance explains the robustness under extreme sparsity.
Moreover, as shown above, the leading eigenvector aligns almost entirely
with the global basis channel $E_0$, indicating that the dominant mode
corresponds to the global scale component.
Since this mode is consistently shared across anchors
and accounts for most of their variation,
the global scale correction can be reliably estimated even from a single anchor.
Additional anchors primarily improve estimation along weaker directions associated with smaller eigenvalues.

% drop anchor robustness
\begin{table}[t]
\centering
\caption{Drop-anchor test: weighted average over KITTI, UrbanSyn, SUN RGB-D (weighted by image count).}
\label{tab:drop_anchor_results}

\renewcommand{\arraystretch}{1.2}
\setlength{\tabcolsep}{10pt}  % default is 6pt
\scalebox{0.8}{
\begin{tabular}{c c c c c}
\toprule
& \multicolumn{2}{c}{DA3} & \multicolumn{2}{c}{MiDaS} \\
\cmidrule(lr){2-3}\cmidrule(lr){4-5}
Anchors
& AbsRel & $\delta_1$
& AbsRel & $\delta_1$ \\
\midrule
1 & 0.18 & 0.74 & 0.16 & 0.78 \\
3 & 0.11 & 0.87 & 0.10 & 0.89 \\
5 & 0.10 & 0.89 & 0.09 & 0.91 \\
7 & 0.10 & 0.90 & 0.09 & 0.92 \\
9 & 0.10 & 0.91 & 0.09 & 0.92 \\
\bottomrule
\end{tabular}}
\end{table}

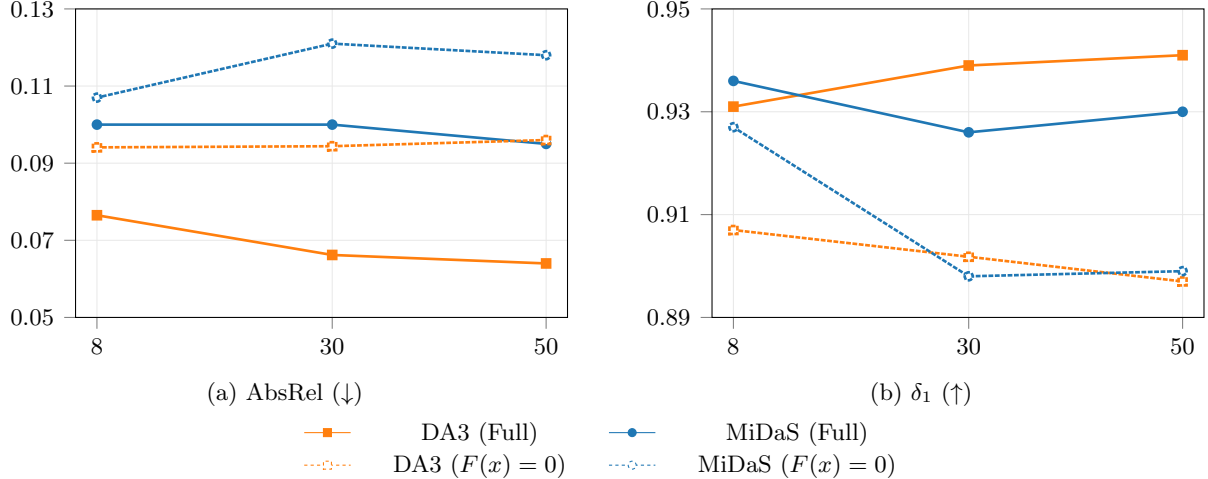
\begin{figure}[t]
\centering
\definecolor{midasC}{RGB}{31,119,180}
\definecolor{da3C}{RGB}{255,127,14}
\definecolor{whitefill}{RGB}{255,255,255}

\pgfplotsset{
  cueAxis/.style={
    width=\linewidth,          % fill *subfigure* width
    height=0.7\linewidth,     % wide & short
    xmin=6, xmax=52,
    xtick={8,30,50},
    tick align=outside,
    tick pos=left,
    grid=major,
    major grid style={draw=gray!18, line width=0.2pt},
    axis line style={line width=0.4pt},
    tick style={line width=0.4pt},
    line width=1pt,
    scaled y ticks=false,
    yticklabel style={
      /pgf/number format/fixed,
      /pgf/number format/precision=2,
      font=\small
    },
    xticklabel style={font=\small},
    xlabel={},
    ylabel={},
    enlarge x limits=false,
    enlarge y limits=false,
    every axis plot/.append style={
      line join=miter,
      line cap=rect,
    },
  }
}

% ================= Left plot =================
\begin{subfigure}[t]{0.49\textwidth}
\centering
\begin{tikzpicture}
\begin{axis}[
  cueAxis,
  ymin=0.05, ymax=0.13,
  ytick={0.05,0.07,0.09,0.11,0.13},
  legend to name=sharedlegend,
  legend style={
    font=\small,
    draw=none,
    fill=none,
    legend columns=2,
    column sep=14pt,
  },
]

% DA3 (full) - square filled
\addplot+[
  color=da3C, solid,
  mark=square*,
  mark size=1.5pt,
  mark options={draw=da3C, fill=da3C, draw opacity=1, fill opacity=1, line join=miter},
] coordinates {(8,0.0765) (30,0.0662) (50,0.064)};
\addlegendentry{DA3 (Full)}

% MiDaS (full) - circle filled
\addplot+[
  color=midasC, solid,
  mark=*,
  mark size=1.5pt,
  mark options={draw=midasC, fill=midasC, draw opacity=1, fill opacity=1},
] coordinates {(8,0.100) (30,0.100) (50,0.095)};
\addlegendentry{MiDaS (Full)}

% DA3 (w/o feat.) - dotted + square filled
\addplot+[
  color=da3C,
  dotted,
  dash pattern=on 0.8pt off 1.6pt,
  mark=square*,
  mark size=1.5pt,
  mark options={draw=da3C, fill=whitefill, draw opacity=1, fill opacity=1, line join=miter},
] coordinates {(8,0.0941) (30,0.0944) (50,0.0960)};
\addlegendentry{DA3 ($F(x)=0$)}

% MiDaS (w/o feat.) - dotted + circle filled
\addplot+[
  color=midasC,
  dotted,
  dash pattern=on 0.8pt off 1.6pt,
  mark=*,
  mark size=1.5pt,
  mark options={draw=midasC, fill=whitefill, draw opacity=1, fill opacity=1},
] coordinates {(8,0.107) (30,0.121) (50,0.118)};
\addlegendentry{MiDaS ($F(x)=0$)}

\end{axis}
\end{tikzpicture}

\caption*{\small (a) AbsRel ($\downarrow$)}
\end{subfigure}
\hfill
% ================= Right plot =================
\begin{subfigure}[t]{0.49\textwidth}
\centering
\begin{tikzpicture}
\begin{axis}[
  cueAxis,
  ymin=0.89, ymax=0.95,
  ytick={0.89,0.91,0.93,0.95},
]

% DA3 (full)
\addplot+[
  color=da3C, solid,
  mark=square*,
  mark size=1.5pt,
  mark options={draw=da3C, fill=da3C, draw opacity=1, fill opacity=1, line join=miter},
] coordinates {(8,0.931) (30,0.939) (50,0.941)};

% MiDaS (full)
\addplot+[
  color=midasC, solid,
  mark=*,
  mark size=1.5pt,
  mark options={draw=midasC, fill=midasC, draw opacity=1, fill opacity=1},
] coordinates {(8,0.936) (30,0.926) (50,0.930)};

% DA3 (w/o feat.)
\addplot+[
  color=da3C,
  dotted,
  dash pattern=on 0.8pt off 1.6pt,
  mark=square*,
  mark size=1.5pt,
  mark options={draw=da3C, fill=whitefill, draw opacity=1, fill opacity=1, line join=miter},
] coordinates {(8,0.907) (30,0.9018) (50,0.897)};

% MiDaS (w/o feat.)
\addplot+[
  color=midasC,
  dotted,
  dash pattern=on 0.8pt off 1.6pt,
  mark=*,
  mark size=1.5pt,
  mark options={draw=midasC, fill=whitefill, draw opacity=1, fill opacity=1},
] coordinates {(8,0.927) (30,0.898) (50,0.899)};

\end{axis}
\end{tikzpicture}

\caption*{\small (b) $\delta_1$ ($\uparrow$)}
\end{subfigure}

\pgfplotslegendfromname{sharedlegend}
\caption{
Results of the ablation study under the medium-anchor regime across 3 datasets (weighted by image count).
Results are shown for DA3 and MiDaS, comparing the full model with a variant where $F(x)=0$ as the number of basis maps (x-axis) increases.
Left: AbsRel ($\downarrow$). Right: $\delta_1$ ($\uparrow$).
Solid lines denote the full model; dotted lines denote the $F(x)=0$ variant.
}
\label{fig:ablation_figure}
\end{figure}

\noindent\textbf{Results of the Ablation Study.}
\Cref{fig:ablation_figure} shows the results of the ablation study under the medium-anchor regime.
As $K$ increases, the full models improve or remain stable. In contrast, the models with $F(x)=0$ degrade with larger $K$.

This behavior suggests that increasing the $K$ is effective only when sufficiently informative cues are provided as input.
In particular, the intermediate representations $F(x)$ offer richer contextual information than depth estimations alone, enabling the model to better utilize additional basis maps.
When $F(x)=0$, increasing $K$ does not yield consistent improvements.
% Ablation: basis number

\section{Discussion}
\textbf{Limitations.}
\label{sec:limitations_future}
We have evaluated our approach on two representative MDE models, DA3~\cite{lin2025depthany3} and MiDaS~\cite{midas,dpt}, demonstrating generalization across modern transformer-based architectures. Validation on additional MDE models, including CNN-based approaches~\cite{monodepth2,dorn,laina2016deeper}, remains future work.

Like other methods, metric recovery ultimately depends on anchor informativeness. Although our formulation improves robustness under sparse anchor regimes, performance may degrade when anchors lack spatial coverage or representation diversity, reflecting an inherent limitation of metric depth recovery.
\textbf{Future Directions.}
The proposed formulation enables several extensions. 
First, it can be incorporated into pseudo-label generation for monocular depth training.
DA3 adopts a teacher–student framework, where a pretrained teacher network produces depth estimations that are aligned to sparse metric measurements and used as supervision for training the student model~\cite{lin2025depthany3}.
Currently, DA3 performs this alignment using a global scale-shift model via RANSAC~\cite{ransac}; replacing this step with the proposed alignment framework may improve pseudo-label accuracy under sparse supervision and reduce inherited bias.

Robust anchor selection strategies that promote representation diversity and spatial coverage represent another natural extension, as better anchor selection may improve metric depth recovery  without increasing anchor density. Finally, the interpretable basis map–weight structure provides a potential diagnostic tool for analyzing bias in pretrained MDE models.

\section{Conclusion}
\label{sec:conclusion}

We propose a structured formulation for metric depth recovery from monocular depth estimation with sparse anchors. The scale field is modeled through a set of image-adaptive basis maps, reducing metric depth recovery to estimating a small set of weights via a least-squares problem, enabling accurate reconstruction even under extreme anchor sparsity.

Across multiple datasets and diverse MDE models, our low-dimensional representation consistently improves metric depth alignment accuracy and maintains robustness across anchor regimes. Beyond quantitative gains, the learned basis maps provide an interpretable decomposition of scale variation, offering a practical framework for incorporating sparse metric information into modern monocular depth estimation systems.

\bibliographystyle{unsrt}
\bibliography{refs}

\end{document}